# Oriented Object Detection in Aerial Images Based on Area Ratio of Parallelogram

Xinyi Yu[a], Mi Lin[a], Jiangping Lu[a], Linlin Ou[a,*]

[a]College of Information Engineering, Zhejiang University of Technology, Hangzhou 310014, China

*Corresponding author. E-mail address: linlinou@zjut.edu.cn (L. Ou).

## Abstract

Oriented object detection is a challenging task in aerial images since the objects in aerial images are displayed in arbitrary directions and are frequently densely packed. The mainstream detectors describe rotating objects using a five-parament or eight-parament representations, which suffer from representation ambiguity for orientated object definition. In this paper, we propose a novel representation method based on area ratio of parallelogram, called ARP. Specifically, ARP regresses the minimum bounding rectangle of the oriented object and three area ratios. Three area ratios include the area ratio of a directed object to the smallest circumscribed rectangle and two parallelograms to the minimum circumscribed rectangle. It simplifies offset learning and eliminates the issue of angular periodicity or label point sequences for oriented objects. To further remedy the confusion issue of nearly horizontal objects, the area ratio between the object and its minimal circumscribed rectangle is employed to guide the selection of horizontal or oriented detection for each object. Moreover, the rotated efficient Intersection over Union (R-EIoU) loss with horizontal bounding box and three area ratios are designed to optimize the bounding box regression for rotating objects. Experimental results on remote sensing datasets, including HRSC2016, DOTA, and UCAS-AOD, show that our method achieves superior detection performance than many state-of-the-art approaches.

## 1. Introduction

Object detection is a vital computer vision technique that aims at classifying and locating the object in images. With the rapid development of satellite remote sensing images and the explosive growth of remote sensing data, object detection in aerial images has been a specific yet activated topic in computer vision [1], [2]. It is widely used in military surveillance, crop monitoring, traffic planning, among other applications. Most traditional methods [3], [4] that rely on manual characteristics to identify objects are sensitive to environmental disruptions such as clouds, sunlight, and rain. Object detection has advanced dramatically in recent years because of deep convolutional neural networks (CNN). Compared with traditional object detection methods, object detection algorithms based on deep learning have the advantages of high accuracy and strong robustness. They are frequently utilized for object detection in remote sensing images.

Object detection can generally be divided into horizontal detection [5], [7], [9], [14] and rotated detection [27], [28], [32], based on the direction of detected boxes. Object detection in aerial images is challenging since the objects have various scales, aspect ratios, and orientations. Consequently, the horizontal object detector for aerial images may lead to misalignment between the detected bounding boxes and ground truth bounding boxes, as shown in Fig. 1(a). The horizontal object detector regresses the horizontal bounding box with four parameters: the abscissa and ordinates of the central point and the length and width of the bounding box. Compared with the horizontal object detector, the rotating object detector adds a regression angle with a total of five parameters. As illustrated in Fig. 1(b), the rotating detector regresses the smallest circumscribed rectangle of objects, successfully resolving the problem of border misalignment and minimizing the overlapping area of dense objects. Therefore, the oriented object detector is preferred for capturing objects in aerial images.

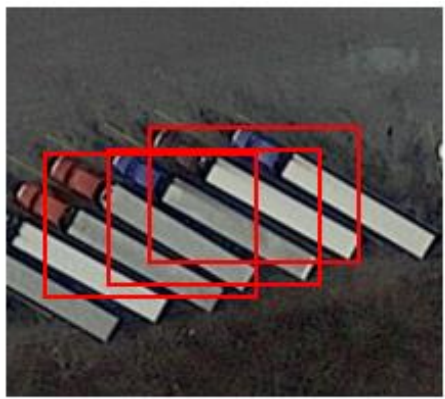 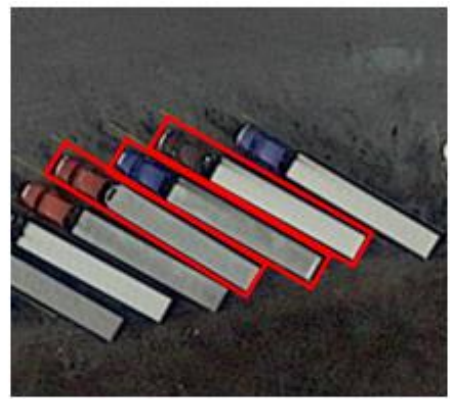

(a) Horizontal object detector      (b) Oriented object detector

Figure 1. Illustration of horizontal object detector and oriented object detector.

Currently, oriented object detectors characterize oriented objects using five-parameter or eight-parameter regression techniques, which introduce representation ambiguity and border discontinuity. Five-parameter regression methods [27], [28] adopt $(x, y, w, h, \theta)$ to represent the rotating objects, which causes periodicity of angle (PoA) and exchangeability of edges (EoE). The predicted angle may be beyond the defined range. Therefore, the model needs to regress the angle with the complicated form at the boundary, which increases the difficulty of predicting the angle and affects the performance of the model. This issue is especially pronounced for oriented objects with wide aspect ratios, such as harbors and bridges in aerial images, as well as long text lines in scene images. Eight-parameter regression methods Eight-parameter regression methods [31] typically utilize four vertices to represent the rotating objects, which creates ambiguity when defining the bounding box order of the four vertices. Theoretically, there are 24 different representations for a single bounding box, so they also face the problem of border discontinuity. As a result, whether employing five-parameter or eight-parameter regression, an object may be represented in various ways. They complicate the regression process and degrade detection performance further.

Several approaches have been proposed to redefine representation for oriented objects, such as BBAVectors [33], Gliding vertex [32]. However, they are regression-based detection approaches, and the underlying boundary problem has not been solved. SCL [34] and DCL [35] are novel approaches to solve boundary issues. The prediction of the object angle is transformed into a classification problem to limit the forecast result to a specified range. Nevertheless, they lost some accuracy and increased the amount of calculation.

In order to solve the non-unique representation of rotating objects, this article suggests a straightforward yet effective representation called ARP. It uses central point coordinates, width and height of the minimal bounding rectangle of the oriented object and introduces three area ratios to represent a rotating object. Three area ratios are as follows: one is the area ratio of the rotating object to its smallest boundary rectangle; another two are the area ratios of two parallelograms to the minimum bounding rectangle of the rotating object (explained in section 3.2). The area ratio can be used to determine the aspect ratio of the corresponding side at the intersection of the oriented object and the horizontal bounding box. It facilitates offset learning and is superior to forecast five-parameter or eight-parameter bounding boxes directly. Continuously, to further eliminate the issue of almost horizontal objects being confused, the area ratio between the object and its minimum circumscribed rectangle is employed to determine whether to detect the object horizontally or oriented. Lastly, a rotation efficient IoU loss (R-EIoU) is suggested to handle the undifferentiable problem and enhance the accuracy of the rotating bounding box. In summary, the main contributions of this paper are:

1. An oriented object representation method (ARP) is proposed based on area ratio of parallelogram. Our method avoids discontinuous regression objects by converting the area ratio into the aspect ratio of the corresponding side at the intersection of the oriented object and the horizontal bounding box. Compared to the baseline methods that learn five-parament or eight-parament, it achieves better performance. The area ratio between the object and its minimum circumscribed rectangle is used to determine whether to detect the object horizontally or oriented, resolving the confusing issue.

2. As an alternative to the smooth L1 loss, we propose a novel loss function called R-EIoU loss. It successfully solves the boundary problem of orientated bounding box regression and achieves higher accuracy in rotated detection when compared to the smooth L1 loss.

3. Experiments show that our method can be easily embedded in different backbones and significantly improve the performance in oriented object detection. It achieves superior performance on four public datasets and three popular detectors.

## 2. Related Work

**2.1 Horizontal Object Detection**

In recent years, object detection has advanced significantly and can be roughly classified into two categories: two-stage detectors and one-stage detectors. The two-stage detectors initially create several areas of interest (RoIs) and then utilize the characteristics of those areas to anticipate object categories and regress the bounding boxes. Numerous high-performance two-stage detectors have been developed, such as R-CNN [5]，Fast R-CNN [6]，Faster R-CNN [7], Mask R-CNN [8]. Although two-stage approaches have shown promising results on some benchmarks, one-stage detectors simplify the detection process by treating it as a regression problem and achieving higher speed. YOLO [9] and its variances (YOLO9000 [10], YOLOv3 [11], YOLOv4 [12], YOLOv5), SSD [13], and RetinaNet [14] are representative single-stage approaches. Compared with anchor-based methods, various anchor-free techniques, such as Cornernet [15] and CenterNet [16], have exploded in popularity in recent years, opening up a new direction for object detection.

Loss function plays a critical role in object detection. It has a significant impact on object detection performance and continues to improve with the advancement of object detectors. Since Smooth L1 loss [6] is robust and resistant to outliers, it is widely utilized in many object detectors such as Fast R-CNN and Faster R-CNN. This approach assumes the four points of the bounding box are independent, whereas they are actually connected. It has been demonstrated that the use of IoU-induced loss[17], such as GIoU[19], DIoU[20], and Focal-EIoU[21], can assure the consistency of the final detection metric and loss in conventional horizontal detectors. However, these IoU losses cannot be used directly in rotating object detection due to the indifferentiability of the rotating IoU.

The above detectors generate bounding boxes solely in the horizontal direction, which frequently mismatch the objects in aerial images due to the arbitrarily rotated and densely distributed objects in remote sensing images. As a result, the recognition of rotatable objects has become a prominent direction in aerial images and scene text.

**2.2 Rotated Object Detection**

Generally, rotated object detectors use oriented bounding boxes to describe the positions of objects, which eliminate the issue of misalignment and are excellent for capturing objects in aerial images and scene text. Numerous effective detectors for scene text detection have been proposed, including $R^2$CNN [22], RRPN [23], TextBoxes++ [24], RRD [25], EAST [26]. These approaches are generally extended from horizontal object detectors by including an angle and some measures to adapt the detection of objects in scene text.

However, object detection is more challenging in aerial images since the objects are frequently small, numerous, and scattered with large-scale changes. It is hard to demonstrate promising performance in remote sensing images using the excellent methods described above. Hence, many robust rotated object detectors have emerged in aerial images. For example, RoI transformer [27] suggests RRoI Leaner and RRoI Warping. RRoI Leaner transforms horizontal RoIs into rotated RoIs, and RRoI Wrapping extracts the rotation-invariant feature from the oriented RoI for subsequent object classification and location regression. $R^3$Det [28] is a one-stage rotation detector based on RetinaNet that solves the feature misalignment problem by employing cascade regression and refining the bounding box. CFC-Net [29] designs a Polarized Attention Module (PAM), refines the preset horizontal anchor points, and adopts Dynamic Anchor Learning (DAL) strategy to achieve high performance in rotating object detection.

While the preceding solutions are promising, they overlook the issue of the discontinuous border. They lead to discontinuous regression loss and impede the convergence of network training. Thus, multiple rotational object detectors have been developed, each of which attempts to overcome the obstacles mentioned above in a unique way. For instance, SCRDet [30] and RSDet [31] propose IoU-smooth L1 loss and modulated loss to alleviate the sharp increase loss caused by the angular periodicity, allowing for improved handling of small, clutter, and rotated objects. Gliding Vertex [32] glides the vertex of the horizontal bounding box to represent a rotatable object, alleviating the issue of border discontinuity and achieving more precise object detection. BBAVectors [33] presents a box boundary-aware bounding box description. BBAVectors is distributed in the four quadrants of the Cartesian coordinate system, which is superior to directly forecasting the spatial characteristics of bounding boxes. CSL [34] and DCL [35] regard angular prediction as a classification task to avoid the discontinuity of regression loss. RIL [36] optimizes the bounding box regression for the rotating objects through Representation Invariance Loss, which solves the inconsistency between the loss and localization quality caused by ambiguous representations. GWD [37] introduces the Gaussian Wasserstein Distance Loss to approximate the non-differentiable rotation IoU-induced loss, which can elegantly resolve border discontinuity and square-like problems.

In summary, all of the solutions above turn definitions of rotating objects or create more precise loss functions to handle the border discontinuity problem. Nonetheless, there is no fully unified answer to all of the issues above. In order to increase the performance and speed in aerial images and scene text, the ARP method is proposed to accurately describe a rotating object and create an R-EIoU loss to approximate the non-differentiable rotation IoU.

## 3. Proposed Method

This section will elaborate on the suggested method in the sequence mentioned below. First, the general architecture of our technique is depicted in Fig. 2. A single-stage rotation detector based on YOLOv5 is utilized in this embodiment. Then, ARP is designed to describe rotating objects accurately, which has a unique representation. To be more precise, the directed object is regressed using its coordinates of the central point, the width and height of its smallest bounding rectangle, and three area ratios. It is easier to learn the orientation and scale information of the object, effectively solving the problem of boundary discontinuity. Additionally, the area ratio between the rotating object and its minimum circumscribed rectangle is also deployed as an obliquity factor for selecting the oriented or horizontal bounding boxes. At last, to more correctly express loss in oriented object identification, the R-EIoU loss is designed.

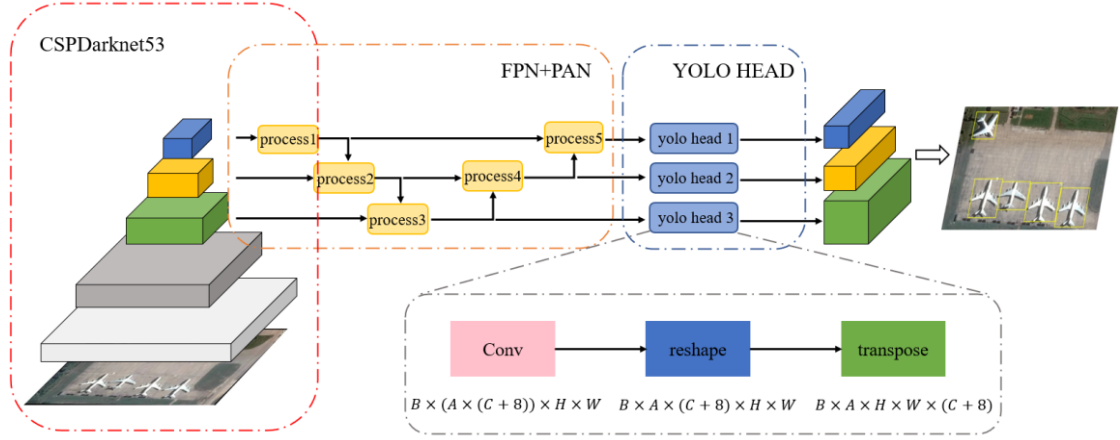

Figure 2. Network architecture of proposed rotation detector (YOLOv5 as an embodiment). CSPDarknet53, FPN+PAN, and YOLO HEAD are the primary components of the proposed detector. We add three extra object variables $(\lambda_1, \lambda_2, \lambda_3)$ to the head of YOLOv5. The symbols 'H' and 'W' refer to the length and width of the grid, and 'B' is the batch size of the sample. 'C' and 'A' represent the number of categories and predefined anchors, respectively, while '8' denotes the confidence and ARP representation method $(x, y, w, h, \lambda_1, \lambda_2, \lambda_3)$.

## 3.1 Network Architecture

YOLOv5 is one of the most advanced single-stage object detectors. We extend YOLOv5 to the oriented object detection to prove the effectiveness of ARP. Its network architecture is sketched in Fig.2. Append three additional object variables $(\lambda_1, \lambda_2, \lambda_3)$ to the head of YOLOv5 and alter its loss function to favor orientated object recognition. Specifically, The CSPDarknet53 is used as the backbone for extracting deep features and generating bounding boxes. The Neck of our approach is FPN and PAN. FPN conveys powerful semantic features from top to bottom, while PAN conveys robust positioning features from bottom to top. Combining FPN and PAN, the parameters of different detection layers are aggregated from different backbone layers. Finally, the category, confidence, and parameters of the direction box are produced via YOLO HEAD, including a horizontal bounding box $(x, y, w, h)$, as well as three area ratios $(\lambda_1, \lambda_2, \lambda_3)$. Additionally, the area ratio $\lambda_1$ between the rotating object and its smallest circumscribed rectangle is employed to determine the obliquity of the orientated or horizontal bounding boxes. If the obliquity factor of each candidate is more than the threshold, suggesting that the underlying object is roughly horizontal, the horizontal bounding box is chosen as the final detection. Otherwise, oriented bounding is used as the final detection method.

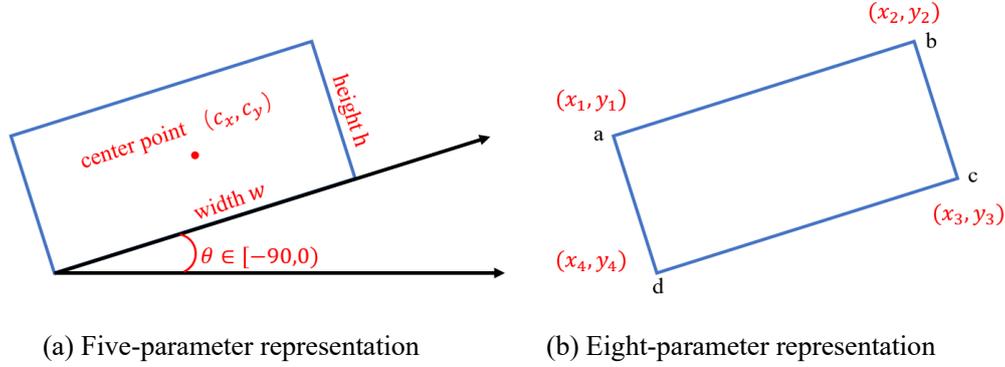

(a) Five-parameter representation    (b) Eight-parameter representation

Figure 3. Common representation method for oriented object detection

**3.2 ARP Representation Method**

3.2.1 Representation Method

Most current object detection systems use five-parameter or eight-parameter regression techniques to describe rotating objects. As illustrated in Fig. 3(a), the five-parameter formulation is consistent with that of OpenCV, which uses the center point coordinates $(C_x, C_y)$, width $w$, height $h$, and angle $\theta$ of the oriented boxes to describe oriented objects. While a tiny angle variation has a negligible effect on the total loss during training, it can result in a substantial IoU disparity between the predicted and ground-truth boxes. Therefore, boundary regression will exhibit inconsistencies. As demonstrated in Fig. 3(b), the eight-parameter method typically uses four boundary points $(x_1, y_1, x_2, y_2, x_3, y_3, x_4, y_4)$ to represent the oriented object. The sequence of the four corner points introduces inconsistencies in the regression process of the model. Thus, whether the five-parameter or eight-parameter regression is applied, border discontinuities arise.

This paper introduces a simple representation for oriented objects and a novel detection scheme that divides and conquers nearly horizontal and oriented object detection. To be more specific, we offer the ARP approach for precisely describing an oriented object parameterized by seven tuples $(x, y, w, h, \lambda_1, \lambda_2, \lambda_3)$, as illustrated in Fig. 4. For every rotating object $P_o$ (see the blue box in Fig. 4), its left, upper, right, and bottom vertices are represented by $(a, b, c, d)$. The smallest circumscribed rectangle of a rotating object is $P_h$ (view the black box in Fig. 4), its four vertices are denoted by $(A, B, C, D)$. $(x, y, w, h)$ denotes the central point, width, and height of the minimal bounding rectangle of the rotating object $P_h$. $\lambda_1$ is the area ratio between the oriented object $P_o$ and minimum circumscribed rectangle $P_h$. $\lambda_2$ is the area ratio of the parallelogram $P_a$ (see the green box in Fig. 4) to the smallest circumscribed rectangle $P_h$. The parallelogram $P_a$ is formed by the extension line of the right side of the top vertex of an oriented object and the extension line of the width of the smallest circumscribed rectangle. $\lambda_3$ is the area ratio of the parallelogram $P_b$ (represented by the red box in Fig. 4) to the smallest circumscribed rectangle $P_h$. The parallelogram $P_b$ comprises the extension line of the right side of the top vertex on the oriented object and the extension line of the height of the smallest circumscribed rectangle. The parameter $\lambda_1$, $\lambda_2$ and $\lambda_3$ are defined as follows:

$$\begin{cases} \lambda_1 = S_o/S_h, \\ \lambda_2 = S_a/S_h, \\ \lambda_3 = S_b/S_h, \end{cases} \quad (1)$$

where $S_o$, $S_h$, $S_a$, and $S_b$ are the areas of the oriented object, minimum circumscribed rectangle, and two parallelograms, respectively.

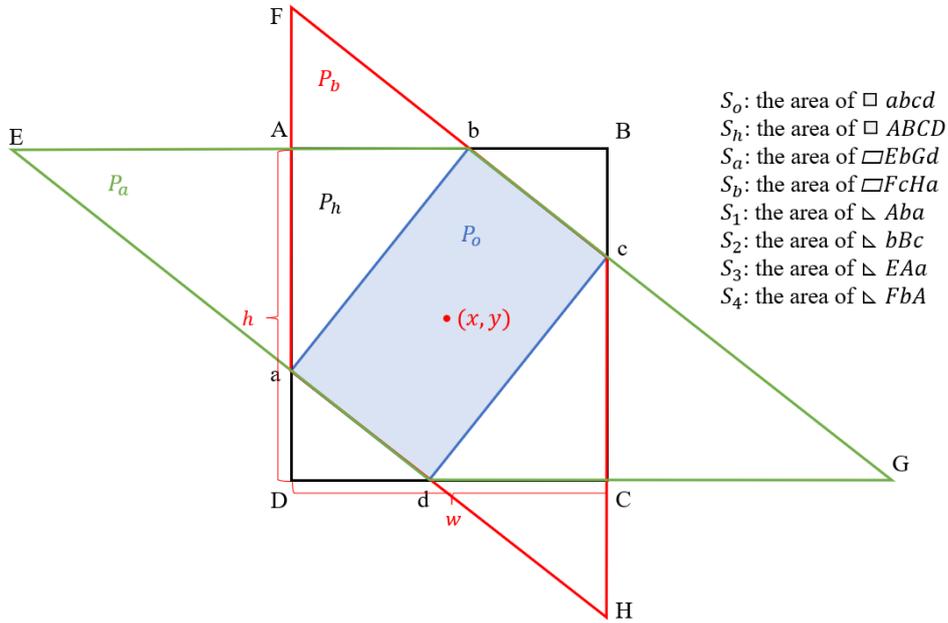

Figure 4. Illustration of ARP representation method. Rectangles colored in blue, black, green, and red represent the oriented object $P_o$, the minimum circumscribed rectangle $P_h$ and two parallelograms $P_a$, $P_b$, respectively, and their areas are $S_o$, $S_h$, $S_a$ and $S_b$. We adopt $(x, y, w, h, \lambda_1, \lambda_2, \lambda_3)$ to represent oriented objects.

3.2.2 Side Lengths and Vertex Coordinates Solution

According to the ARP representation method in Fig. 4, the area of the minimum circumscribed and two parallelograms can be represented by the area of oriented object and triangles as follows:

$$\begin{cases} S_h = S_o + 2(S_1 + S_2), \\ S_a = S_o + 2(S_1 + S_3), \\ S_b = S_o + 2(S_1 + S_4), \end{cases} \quad (2)$$

where $S_1$, $S_2$, $S_3$, and $S_4$ denote the areas of triangles $\varDelta Aba$, $\varDelta bBc$, $\varDelta EAa$, and $\varDelta FbA$, respectively. Since the oriented object in aerial images is denoted by a rectangular box, and its smallest circumscribed rectangle is also a rectangle box, our represent method contains two pairs of similar triangles $\varDelta EAa \sim \varDelta bBc$ and $\varDelta FbA \sim \varDelta bBc$. Given their similarity ratios of $k_1$ and $k_2$, respectively. We can derive the following relationships between the four triangles:

$$\begin{cases} S_3/S_2 = k_1^2, \\ S_4/S_2 = k_2^2, \\ S_1/S_2 = k_1 k_2. \end{cases} \quad (3)$$

Combine Eqs. (2) and (3), the values of $k_1$ and $k_2$ can be obtained as follows:

$$\begin{cases} k_1 = \sqrt{\frac{(S_a - S_o)^2}{(S_h - S_o)(S_a - S_o) + (S_h - S_a)(S_b - S_o)}}, \\ k_2 = \sqrt{\frac{(S_b - S_o)^2}{(S_h - S_o)(S_a - S_o) + (S_h - S_a)(S_b - S_o)}}. \end{cases} \quad (4)$$

By substituting Eq. (4) into Eq. (1), $k_1$ and $k_2$ represented by the area ratios are as follows:

$$\begin{cases} k_1 = \sqrt{\frac{(\lambda_2 - \lambda_1)^2}{(1-\lambda_1)(\lambda_2 - \lambda_1) + (1-\lambda_2)(\lambda_3 - \lambda_1)}}, \\ k_2 = \sqrt{\frac{(\lambda_3 - \lambda_1)^2}{(1-\lambda_1)(\lambda_2 - \lambda_1) + (1-\lambda_2)(\lambda_3 - \lambda_1)}}. \end{cases} \quad (5)$$

Furthermore, the length and coordinates of the intersection point of a rotating object with its smallest circumscribed rectangle can also be acquired as follows:

$$\begin{cases} h_1 = |Aa| = \frac{k_1}{1+k_1} h, \\ h_2 = |aD| = \frac{1}{1+k_1} h, \\ w_1 = |Ab| = \frac{k_2}{1+k_2} w, \\ w_2 = |bB| = \frac{1}{1+k_2} w, \end{cases} \quad (6)$$

$$\begin{cases} a: (x - w/2, y - h/2 + h_1), \\ b: (x - w/2 + w_1, y - h/2), \\ c: (x + w/2, y + h/2 - h_1), \\ d: (x + w/2 - w_1, y + h/2), \end{cases} \quad (7)$$

where $h_1, h_2$ denote the length of line segment Aa and aD, and $w_1, w_2$ indicate the length of the Ab and bB line segment. $(a, b, c, d)$ represent the coordinates of the four vertices of the oriented object.

It is seen that the ARP representation method can accurately describe the position information of the rotating object in the image. We can convert the ARP to the more common eight-parament representation method by applying the Eqs. (5), (6) and (7). Similarly, converting the eight-parameter representation method to our representation method is straightforward.

Thus, the ARP method accurately describes rotating objects and has a unique representation. It solves the discontinuous boundary problem effectively and significantly improves the detection accuracy of rotating objects.

3.2.3 Nearly Horizontal Bounding Box

In practice, we observe that the detection fails when an object extremely approaches the horizontal bounding box, as illustrated in Fig. 5. When the oriented object is close to the horizontal bounding box, the parameter $\lambda_1$ and one of the parameters in $\lambda_2$ and $\lambda_3$ approach one infinitely. Thus, it will be impossible to obtain four vertex coordinates of the rotating object since the value of the parameter $k_1$ or $k_2$ is infinite (calculated by Eq. (5)). This problem is regarded as an extreme case.

To address this case, we divide the rotating objects into oriented bounding boxes (OBBs) and horizontal bounding boxes (HBBs) and process them independently in this work. The advantage of this classification strategy is that it allows us to convert difficult-to-handle corner cases into HBBs cases. A nearly horizontal object has a high area ratio $\lambda_1$ that is close to one (see Fig. 5), whereas a highly slender oriented object has a low area ratio $\lambda_1$ that is close to zero. Therefore, the area ratio $\lambda_1$ are adopted as an obliquity factor to guide the selection of OBB and HBB. Without introducing other parameters, the categories of OBB and HBB can be finished. Indeed, it is reasonable to represent nearly horizontal objects with horizontal bounding boxes.

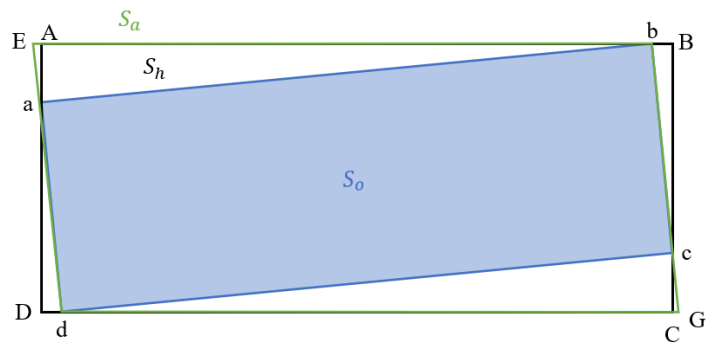

Figure 5: Illustration of the nearly horizontal bounding box cases. The object box $S_o$, the minimum circumscribed rectangle $S_h$ and parallelogram $S_a$ are nearly equal in area.

### 3.3 R-EIoU Loss

The proposed method is based on YOLOv5. Thus the loss function makes some appropriate modifications based on YOLOv5. The multi-task loss is defined as follows:

$$L = \lambda_{box} \sum_{i=0}^{K \times K} \sum_{j=0}^{M} I_{ij}^{obj} L_{box} + \lambda_{obj} \sum_{i=0}^{K \times K} \sum_{j=0}^{M} I_{ij}^{obj} L_{obj} +$$

$$\lambda_{cls} \sum_{i=0}^{K \times K} I_{ij}^{obj} \sum_{c \in cls} L_{cls} + \lambda_\alpha \sum_{i=0}^{K \times K} \sum_{j=0}^{M} I_{ij}^{obj} L_\alpha, \tag{8}$$

where $K \times K$ denotes the number of grids, $M$ indicates the number of candidate boxes produced by each grid. Each candidate box will get a corresponding bounding box through the network, so we finally obtain the $K \times K \times M$ bounding boxes. $I_{ij}^{obj}$ is a binary label that indicates whether the $j$ anchor box of the $i$ grid is responsible for detecting this object. If it is responsible, it is set to 1; otherwise, it is set to 0. The hyper-parameters $\lambda_{obj}$, $\lambda_{cls}$, $\lambda_{box}$, and $\lambda_\alpha$ are used to balance the importance of each loss term. The confidence loss $L_{obj}$ and classification loss $L_{cls}$ are both Focal losses, which are the same as that in YOLOv5. $L_{box}$ signifies the loss for bounding box regression. $L_\alpha$ denotes the loss for obliquity factor $\lambda_1$ regression, which is defined as the binary cross-entropy:

$$L_{cls} = -(\alpha \cdot \log(p(\alpha)) + (1 - \alpha) \log(1 - p(\alpha))), \tag{9}$$

where $\alpha$ denotes a binary label of either 0 or 1, $p(\alpha)$ represents the probability that the output corresponds to the $\alpha$ label. The following is the definition of $\alpha$:

$$\alpha = \begin{cases} 1 \ (OBB) & \lambda_1 < \lambda_{thr}, \\ 0 \ (HBB) & otherwise, \end{cases} \tag{10}$$

where $\lambda_{thr}$ is the threshold used to differentiate between the oriented and horizontal bounding boxes, and its optimal value will be determined in section 4.2.3. The loss function for the $L_{box}$ regression contains two terms for the horizontal bounding box $(x, y, w, h)$ and three area ratios $(\lambda_1, \lambda_2, \lambda_3)$. In terms of the horizontal bounding box, $L_{box}$ is the same as the CIoU loss in YOLOv5. Compared with the horizontal bounding, the oriented bounding box adds three parameters $\lambda_1, \lambda_2, \lambda_3$ in the head. Therefore, it needs to augment extra loss to calculate these three variables and facilitate their regression.

Smooth L1 loss is a relatively excellent loss function in rotating object detection, which can promote the regression of the object without the problem of non-differentiable. Therefore, we also considered using smooth L1 loss to initially deal with the parameters of these three area ratios. Given a predicted horizontal bounding box $B$ and an object horizontal bounding box $\hat{B}$, their losses are defined as follows:

$$L_{box} = L_{CIoU} + \sum_{i=1}^{3} smoothL_1(\lambda_i - \hat{\lambda}_i), \qquad (11)$$

where $L_{CIoU}$ denotes the CIoU loss between the two horizontal boxes $B$ and $\hat{B}$. However, using CIoU to address horizontal bounding and smooth L1 loss to address three area ratios generates boundary problems in the case of nearly horizontal bounding boxes. As demonstrated in Fig. 6(a), some bounding boxes do not correspond to the objects when smooth L1 loss is applied. Therefore, we propose R-EIoU loss to better solve this problem. Given two predicted parallelogram boxes $P_a$ and $P_b$ and two object boxes $\hat{P}_a$ and $\hat{P}_b$ (see Fig. 4), the R-EIoU loss is defined as follows:

$$L_{box} = 1 - IoU + \frac{\rho^2(b,\hat{b})}{c^2} + \frac{\rho^2(w_{pa},\hat{w}_{pa})}{c_{wpa}^2} + \frac{\rho^2(h_{pb},\hat{h}_{pb})}{c_{hpb}^2}, \qquad (12)$$

where $IoU$ represent the Intersection over Union between the $B$ and $\hat{B}$. $b$ and $\hat{b}$ denote the central points of $B$ and $\hat{B}$, respectively. $\rho(\cdot) = \|\cdot\|$ indicates the Euclidean distance. The diagonal length of the smallest enclosing box that encompasses the two horizontal bounding boxes $B$ and $\hat{B}$ is denoted by $c$. $w_{pa}$ and $\hat{w}_{pa}$ signify the width of $P_a$ and $\hat{P}_a$, respectively. $c_{wpa}$ is the width of the smallest enclosing box covering the two parallelogram boxes $P_a$ and $\hat{P}_a$. $h_{pb}$ and $\hat{h}_{pb}$ denote the height of $P_b$ and $\hat{P}_b$, respectively. The height of the minimum enclosing box covering the two parallelogram boxes $P_b$ and $\hat{P}_b$ is denoted by $c_{hpb}$. Namely, we divide the loss function into three parts: the IoU loss $1 - IoU$, the distance loss $\rho^2(b,\hat{b})/c^2$, and the area ratio loss $\rho^2(w_{pa},\hat{w}_{pa})/c_{wpa}^2 + \rho^2(h_{pb},\hat{h}_{pb})/c_{hpb}^2$.

The R-EIoU loss effectively solves the problem of inaccurate expression of box regression loss function, especially in the case of an approximate horizontal bounding box (see Fig. 6(b)). Experiments show that the detector based on this R-EIoU loss can achieve more considerable gains than smooth L1 loss.

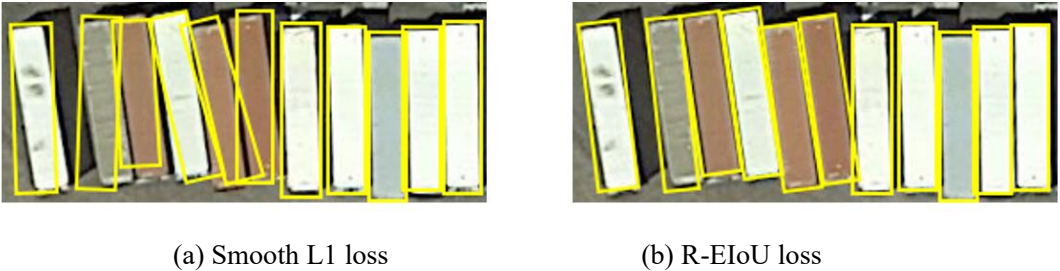

(a) Smooth L1 loss        (b) R-EIoU loss

Figure 6. Comparison of detection results by two losses.

# 4. Experiments

Experiments are performed with Pytorch on a server with 8 GeForce RTX 2080 Ti and $11 \times 8G$ memory. We first give a description of the datasets and then use these datasets to verify the advantages of the proposed method.

## 4.1 Datasets and Implementation Details

### 4.1.1 DOTA Dataset

DOTA[1] is a large-scale benchmark and challenge for object detection in aerial images, and it is currently the most comprehensive remote sensing image data set for object detection. There are three versions of this data set that has been released, including DOTAv1.0, DOTAv1.5, and DOTAv2.0. Since the current rotation object detection algorithms are based on DOTAv1.0, the experiment in this paper is mainly completed on DOTAv1.0 to verify the performance of our proposed method. DOTAv1.0 includes 2,806 aerial images with 188,282 annotated instances, and the size of the image ranges from around $800 \times 800$ to $4,000 \times 4,000$ pixels. There are 15 categories in total, including plane (PL), baseball diamond (BD), bridge (BR), ground track field (GTF), small vehicle (SV), large vehicle (LV), ship (SH), tennis court (TC), basketball court (BC), storage tank (ST), soccer ball field (SBF), roundabout (RA), harbor (HA), swimming pool (SP) and helicopter (HC). It uses 1/2 of the images as the training set, 1/6 as the verification set, and 1/3 as the test set. The test set results are not public, and it needs to be uploaded to the server for the test. Since the ariel images in the DOTA dataset are of varying sizes, we divide the images into $1,024 \times 1,024$ pixel sub-images with an overlap of 200 pixels, with is the same as RoI Transformer [27].

### 4.1.2 HRSC2016 Dataset

HRSC2016 [2] is a challenging dataset for ship detection in aerial images. All images come from six famous ports, including two scenes of marine vessels and offshore vessels. It contains 1,061 images and more than 20 categories of ships in various appearances. The image size ranges from $300 \times 300$ to $1500 \times 900$. The training, validation and test sets contain 436 images (including 1,207 instances), 181 images (including 541 instances) and 444 images (including 1,228 instances).

### 4.1.3 UCAS-AOD Dataset

USCA-AOD [39] is a remote sensing data set for the detection of vehicles and planes issued by the University of Chinese Academy of Sciences. It contains 1,510 aerial images of approximately 659 × 1,280 pixels, with two categories of 14,596 instances in total. In line with ICN [41] and DOTA [1], we randomly select 1,110 images for training and 400 images for testing.

4.1.4 ICDAR2015 Dataset

ICDAR2015 [40] is a data set for scene text detection and location proposed by ICDAR 2015 Robust Reading Competition. It contains a total of 1,500 images, of which 1,000 are utilized for training and the remainder for testing. Text regions are annotated by the four vertices of the quadrangle.

4.1.5 Baseline and Training Details:

To demonstrate the generality of our system, we conduct experiments on three aerial benchmarks and one scene text benchmark. RetinaNet [14], R³Det [28], and YOLOv5 are adopted as the baseline in all ablation studies and adopt YOLOv5 as final experiments. The initial learning rates for RetinaNet, R³Det, and YOLOv5 are 0.0005, 0.001, and 0.01, respectively. The momentum and weight decay are 0.9 and 0.0001, respectively.

The baseline of our experiment is YOLOv5 unless otherwise specified. The training epoch for DOTAv1.0, HRSC2016, USAS-AOD, and ICDIR2015 are 300, 200, 150, and 150, respectively. Additionally, the number of iterations of each epoch is proportional to the sample size in the dataset. The warm-up approach determines an appropriate learning rate during the first quarter of the training epochs. The final detection results are post-processed using rotating non-maximum suppression (R-NMS) during the inference. The hyperparameters $\lambda_{box}$, $\lambda_{cls}$, $\lambda_{obj}$ and $\lambda_{\alpha}$ in Eq. (8) are set to 0.05, 0.3, 0.7 and 0.8, respectively. Without the extra declaration, other hyper-parameter settings related to the program are consistent with YOLOv5.

**4.2 Ablation Study**

In this section, we conduct a series of ablation experiments on DOTA-v1.0, HRSC2016, UCAS-AOD, and ICDAR2015 datasets to evaluate the effectiveness of our proposed method. For fair compassion, all baseline methods are implemented using similar settings to the proposed method. Please take note that sophisticated data augmentation techniques are not employed in this section.

4.2.1 Evaluation of ARP Representation Method

The RetinaNet, R³Det, and YOLOv5 are employed as our baseline to evaluate the effectiveness of the ARP representation method. Notably, except for the output box representations, the baseline method shares the same architecture and loss (smooth L1 loss) as the proposed method. We compare our ARP method to others that employ different representations, as illustrated in Tab. 1. The suggested method outperforms OpenCV representation (DOC) on RetinaNet, R$^3$Det, and YOLOv5 by 3.03%, 0.96%, and 1.71%, respectively, on the DOTAv1.0 dataset. Additionally, comparative tests are conducted on the HRSC2016 dataset to confirm the boosting effect of ARP. Results show that our ARP method also achieves predominant performance. The result suggests that the ARP representation approach is superior to learning the five-parameter (DOC and DLE [37]) and angle classification methods (CSL [34] and DCL [35]) for detecting orientated objects.

Table 1. Ablation study for ARP on three baselines. Where DOC denotes the defined method based on OpenCV, DLE indicates the defined method based on the long edge.

| Method | Backbone | Datasets | BOX DEF | mAP |
| --- | --- | --- | --- | --- |
| RetinaNet | Resnet50 | DOTAv1.0 | DLE | 64.17 |
| | | | DOC | 65.73 |
| | | | CSL | 67.38 |
| | | | DCL | 67.39 |
| | | | **ARP (ours)** | **68.76** |
| | | HRSC2016 | DOC | 84.28 |
| | | | **ARP (ours)** | **86.43** |
| R$^3$Det | Resnet50 | DOTAv1.0 | DOC | 70.66 |
| | | | DCL | 71.21 |
| | | | **ARP (ours)** | **71.62** |
| | | HRSC2016 | DOC | 88.52 |
| | | | **ARP (ours)** | **88.93** |
| YOLOv5 | CSPDarknet53 (yolov5x) | DOTAv1.0 | DOC | 71.12 |
| | | | **ARP (ours)** | **72.83** |
| | | HRSC2016 | DOC | 88.65 |
| | | | **ARP (ours)** | **89.07** |

4.2.2 Evaluation of R-EIoU Loss

The R-EIoU loss for our ARP approach is constructed in the manner described in section 3.3. As illustrated in Tab.2, four distinct datasets are utilized to test the effectiveness of the R-EIoU loss. As mentioned previously, the basic model is YOLOv5. The regression head produces the orientated bounding box, which is represented by four bounding box parameters and three area ratios. R-EIoU loss achieves mAP values of 73.33%, 89.46%, 96.58%, and 84.43% on the DOTAv1.0, HRSC2016, USAS-AOD, and ICDIR2015, respectively, which is an improvement of 0.50%, 0.39%, 0.42%, and 0.27% over the baseline model. R-EIoU loss of the orientated bounding box combines horizontal bounding box and area ratio, resolving boundary difficulties and promoting network convergence. Thus, the R-EIoU loss design is effective for our ARP approach.

Table 2. Ablation study for R-EIoU loss on three datasets.

| Method | Backbone | Datasets | Loss | mAP |
|---|---|---|---|---|
| YOLOv5 | CSPDarknet53 (yolov5x) | DOTAv1.0 | Smooth L1 loss | 72.83 |
| | | | **R-EIoU loss** | **73.33** |
| | | HRSC2016 | Smooth L1 loss | 89.07 |
| | | | **R-EIoU loss** | **89.46** |
| | | UCAS-AOD | Smooth L1 loss | 96.16 |
| | | | **R-EIoU loss** | **96.58** |
| | | ICDAR2015 | Smooth L1 loss | 84.16 |
| | | | **R-EIoU loss** | **84.43** |

4.2.3 Hyperparameters of ARP

The parameter sensitivity experiments are carried out in order to establish the appropriate hyperparameter settings. As discussed in section 3.2.3, ARP regression is accurate for oriented objects but less accurate for objects close to the level due to the possibility of confusion. YOLOv5 is used as a baseline to examine the influence of various obliquity factor thresholds $\lambda_{thr}$ on four datasets. As depicted in Fig. 7, the best thresholds in DOTAv1.0, HRSC2016, USAS-AOD and ICDIR2015 are 0.94 (0.95), 0.92, 0.96 and 0.91 (0.92), respectively. Indeed, when a low threshold is applied, horizontal bounding boxes represent some oriented objects, resulting in lower oriented object identification performance. However, when a wide threshold is employed, almost horizontal objects may have difficulty discriminating between horizontal and orientated bounding boxes. Therefore, it is vital for us to choose appropriate threshold values across distinct datasets.

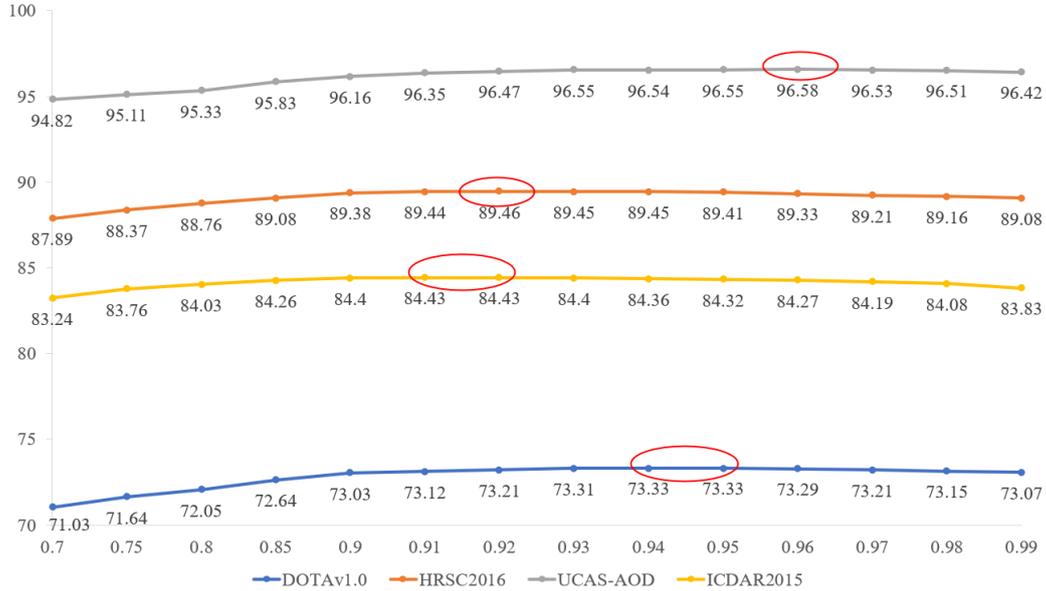

Figure 7. Ablation study on four datasets with varying obliquity thresholds $\lambda_{thr}$.

**4.3 Comparison With State-of-art Methods**

We compare our proposed method to other state-of-the-art methods in three aerial image detection datasets, DOTA, HRSC2016, and UCAS-AOD, as well as one scene text detection dataset, ICDAR2015.

4.3.1 Results on DOTA

Tab. 3 compares the proposed approach to various outstanding detectors on DOTA dataset. Because ground-truth labels of the test set are not publicly available, the detection results should be submitted to the official DOTA assessment site. Our approach achieves 75.31% and 76.05% mAP, respectively, using YOLOv5x and YOLOv5x6 as the backbone. Moreover, As can be observed, introducing R-EIoU improves performance by 1.69% and results in 77.74% mAP. In addition, it achieves 79.93% mAP with YOLOv5x6 in the multi-scale (0.5, 1.0, 1.5, 2.0) during training and testing. The experimental evidence indicates that our method outperforms previous methods, particularly for objects with dense packing and many directions (see SV, LV, SH, and TC in Tab. 3). Fig. 8 illustrates some DOTA dataset results.

Table 3. Performance evaluation of OBB task on DOTAv1.0 dataset. R-101 indicates ResNet-101(likewise for R-50), and H-104 stands for Hourglass-104. YOLOv5x and YOLOv5x6 denote different models in YOLOv5. * denotes multi-scale training and testing.

| Methods | Backbone | PL | BD | BR | GTF | SV | LV | SH | TC | BC | ST | SBF | RA | HA | SP | HC | mAP |
|---|---|---|---|---|---|---|---|---|---|---|---|---|---|---|---|---|---|
| Two-stage: | | | | | | | | | | | | | | | | | |
| RRPN[23] | R-101 | 88.52 | 71.20 | 31.66 | 59.30 | 51.85 | 56.19 | 57.25 | 90.81 | 72.84 | 67.38 | 56.69 | 52.84 | 53.08 | 51.94 | 53.58 | 61.01 |
| ICN[41] | R-101 | 81.36 | 74.30 | 47.70 | 70.32 | 64.89 | 67.82 | 69.98 | 90.76 | 79.06 | 78.20 | 53.64 | 62.90 | 67.02 | 64.17 | 50.23 | 68.16 |
| RoI Trans.[28] | R-101 | 88.64 | 78.52 | 43.44 | 75.92 | 68.81 | 73.68 | 83.59 | 90.74 | 77.27 | 81.46 | 58.39 | 53.54 | 62.83 | 58.93 | 47.67 | 69.56 |
| CAD-Net[42] | R-101 | 87.80 | 82.40 | 49.40 | 73.50 | 71.10 | 63.50 | 76.70 | 90.90 | 79.20 | 73.30 | 48.40 | 60.90 | 62.00 | 67.00 | 62.20 | 69.90 |
| SCRDet[30] | R-101 | 89.98 | 80.65 | 52.09 | 68.36 | 68.36 | 60.32 | 72.41 | 90.85 | **87.94** | 86.86 | 65.02 | 66.68 | 66.25 | 68.24 | 65.21 | 72.61 |
| Gliding Vet.[33] | R-101 | 89.64 | 85.00 | 52.26 | 77.34 | 73.01 | 73.14 | 86.82 | 90.74 | 79.02 | 86.81 | 59.55 | **70.91** | 72.94 | 70.86 | 57.32 | 75.02 |
| MASKOBB[43] | RX-101 | 89.56 | 85.95 | 54.21 | 72.90 | 76.52 | 74.16 | 85.63 | 89.85 | 83.81 | 86.48 | 54.89 | 69.64 | 73.94 | 69.09 | 63.32 | 75.33 |
| CSL[34] | R-152 | **90.25** | 85.53 | 54.64 | 75.31 | 70.44 | 73.51 | 77.62 | 90.84 | 86.15 | 86.69 | 69.60 | 68.04 | 73.83 | 71.10 | 68.93 | 76.17 |
| SCRDet++[44] | R-101 | 90.05 | 84.39 | 55.44 | 73.99 | 77.54 | 71.11 | 86.05 | 90.67 | 87.32 | 87.08 | 69.62 | 68.90 | 73.74 | 71.29 | 65.08 | 76.81 |
| Single-stage: | | | | | | | | | | | | | | | | | |
| DRN[45] | H-104 | 88.91 | 80.22 | 43.52 | 63.35 | 73.48 | 70.69 | 84.94 | 90.14 | 83.85 | 84.11 | 50.12 | 58.41 | 67.62 | 68.60 | 52.50 | 70.70 |
| DAL[46] | R-101 | 88.61 | 79.69 | 46.27 | 70.37 | 65.89 | 76.10 | 78.53 | 90.84 | 79.98 | 78.41 | 58.71 | 62.02 | 69.23 | 71.32 | 60.65 | 71.78 |
| RSDet[31] | R-101 | 89.90 | 82.90 | 48.60 | 65.20 | 69.50 | 70.10 | 70.20 | 90.50 | 85.60 | 83.40 | 62.50 | 63.90 | 65.60 | 67.20 | 68.00 | 72.20 |
| P-RSDet[38] | R-101 | 88.58 | 77.83 | 50.44 | 69.29 | 71.10 | 75.79 | 78.66 | 90.88 | 80.10 | 81.71 | 57.92 | 63.03 | 66.30 | 69.77 | 63.13 | 72.30 |
| RIDet-Q[36] | R-101 | 87.38 | 75.64 | 44.75 | 70.32 | 77.87 | 79.43 | 87.43 | 90.72 | 81.16 | 82.52 | 59.36 | 63.63 | 68.11 | 71.94 | 51.42 | 72.78 |
| CFC-Net[29] | R-50 | 89.08 | 80.41 | 52.41 | 70.02 | 76.28 | 78.11 | 87.21 | 90.89 | 84.47 | 85.64 | 60.51 | 61.52 | 67.82 | 68.02 | 50.09 | 73.50 |
| RIDet-O[36] | R-101 | 88.94 | 78.45 | 46.87 | 72.63 | 77.63 | 80.68 | 88.18 | 90.55 | 81.33 | 83.61 | 64.85 | 63.72 | 73.09 | 73.13 | 56.87 | 74.70 |
| BBAVector[33] | R-101 | 88.63 | 84.06 | 52.13 | 69.56 | 78.26 | 80.40 | 88.06 | 90.87 | 87.23 | 86.39 | 56.11 | 65.62 | 67.10 | 72.08 | 63.96 | 75.36 |
| R$^3$Det[28] | R-152 | 89.80 | 83.77 | 48.11 | 66.77 | 78.76 | 83.27 | 87.84 | 90.82 | 85.38 | 85.51 | 65.67 | 62.68 | 67.53 | 78.56 | 72.62 | 76.47 |
| PolarDet[47] | R-101 | 89.65 | **87.07** | 48.14 | 70.97 | 78.53 | 80.34 | 87.45 | 90.76 | 85.63 | 86.87 | 61.64 | 70.32 | 71.92 | 73.09 | 67.15 | 76.64 |
| R3Det-DCL[35] | R-152 | 89.26 | 83.60 | 53.54 | 72.76 | 79.04 | 82.56 | 87.31 | 90.67 | 86.59 | 86.98 | 67.49 | 66.88 | 73.29 | 70.56 | 69.99 | 77.37 |
| S$^2$A-Net[48] | R-50 | 89.89 | 83.60 | 57.74 | **81.95** | **79.94** | 83.19 | 89.11 | 90.78 | 84.87 | 87.81 | **70.30** | 68.25 | 78.30 | 77.01 | 69.58 | 79.42 |
| ROSD[49] | R-101 | 88.28 | 85.74 | **59.79** | 77.46 | 79.48 | 84.02 | 88.32 | 90.82 | 87.45 | 85.65 | 66.80 | 66.68 | **78.78** | **82.52** | **74.70** | 79.76 |
| **ARP** | YOLOv5x | 85.19 | 83.47 | 51.36 | 75.32 | 73.12 | 78.49 | 85.16 | 90.45 | 80.19 | 86.39 | 59.45 | 64.83 | 76.92 | 77.35 | 61.94 | 75.31 |
| **ARP** | YOLOv5x6 | 85.97 | 83.04 | 52.73 | 76.11 | 75.92 | 79.50 | 88.91 | 90.68 | 79.15 | **88.33** | 61.20 | 63.28 | 77.29 | 75.83 | 62.94 | 76.05 |
| **ARP+R-EIoU** | YOLOv5x6 | 88.53 | 83.71 | 53.76 | 80.07 | 77.42 | 82.19 | 84.70 | 90.79 | 86.73 | 87.44 | 64.04 | 63.39 | 76.77 | 80.67 | 66.03 | 77.74 |
| **ARP+R-EIoU*** | YOLOv5x6 | 89.32 | 85.28 | 58.12 | 81.12 | 78.54 | **84.11** | **89.43** | **90.91** | 87.42 | 87.88 | 68.76 | 67.01 | 77.87 | 81.87 | 71.32 | **79.93** |

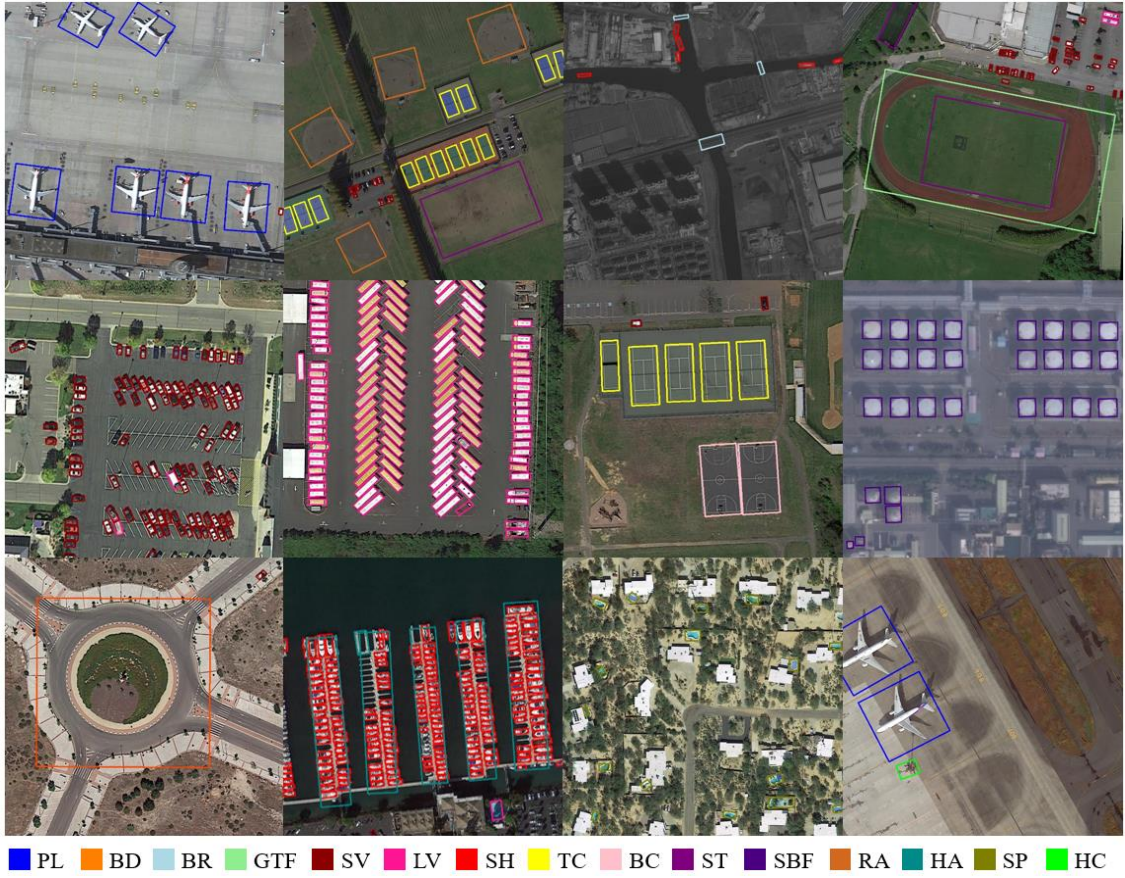

Figure 8. Visualization of the detection results of ARP+R-EIoU on DOTA dataset.

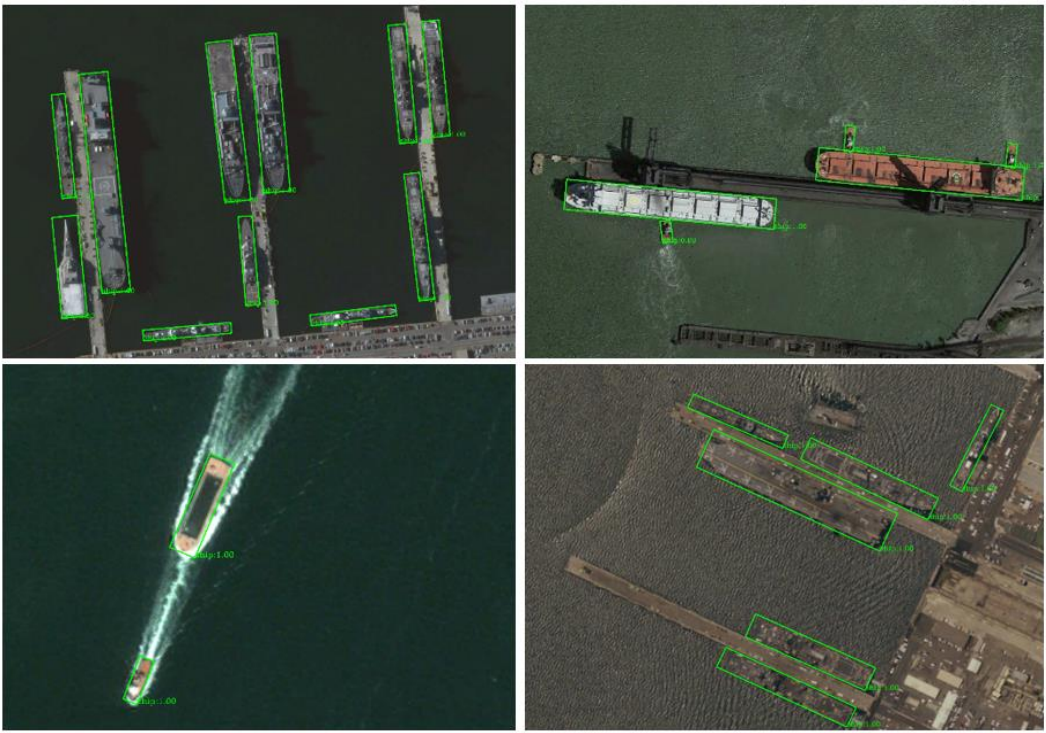

Figure 9. Detection results on HRSC2016 dataset with our method.

Table 4. Results and speed comparison on the HRSC2016 dataset.

| Method | Backbone | mAP (07) | mAP (12) | FPS |
| --- | --- | --- | --- | --- |
| R$^2$CNN [22] | ResNet101 | 73.07 | 79.73 | 2 |
| RRPN [23] | ResNet101 | 79.08 | 85.64 | 3.5 |
| ROI-Transformer [27] | ResNet101 | 86.20 | — | 6 |
| RSDet [31] | ResNet50 | 86.50 | — | — |
| Gliding Vertex [32] | ResNet101 | 88.20 | — | 10 |
| BBAVectors [33] | ResNet101 | 88.60 | — | 11.7 |
| DAL [46] | ResNet101 | 88.95 | — | **34** |
| R$^3$Det [28] | ResNet101 | 89.26 | 96.01 | 12 |
| R$^3$Det-DCL [35] | ResNet101 | 89.46 | 96.41 | — |
| R$^4$Det [17] | ResNet101 | 89.56 | — | 6.5 |
| CSL [34] | ResNet50 | 89.62 | 96.10 | — |
| RIDet-O [36] | ResNet101 | 89.63 | — | — |
| CFC-Net [29] | ResNet101 | 89.70 | — | 28 |
| S$^2$A-Net [48] | ResNet101 | 90.17 | 95.01 | 12.7 |
| PolarDet [47] | ResNet50 | 90.46 | — | 25 |
| **ARP+R-EIoU (Ours)** | YOLOv5x | 89.86 | 96.27 | 30 |
| **ARP+R-EIoU (Ours)** | YOLOv5x6 | **90.52** | **96.63** | 14 |

4.3.2 Results on HRSC2016

The experimental results on HRSC2016 are evaluated under PASCAL VOC2007 and VOC2012 metrics. HRSC2016 dataset contains lots of large aspect ratios and multiple directions ship, which poses a considerable challenge for oriented object detection. As shown in Tab. 4, our method achieves 89.86% and 96.27% mAP under VOC2007 and VOC2012 metrics, respectively, outperforming most methods. Additionally, our model with YOLOv5x reaches 30 FPS on RTX 2080 Ti GPU, exceeding major of detectors. More crucially, our model with YOLOv5x6 delivers state-of-the-art performance, scoring around 90.52% and 96.63% on 2007 and 2012 evaluation metrics, respectively. Fig. 9 illustrates some visualization findings on HRSC2016.

4.3.3 Results on UCAS-AOD

VOC2012 measures are used to evaluate the experimental results on USAC-AOD dataset. As illustrated in Tab. 5, Our approach yields 97.04% mAP in testing, outperforming existing advanced rotation detectors such as R$^3$Det [28] and PolarDet [47]. Furthermore, our technique has the best performance in detecting small cars, indicating that it is robust to small and densely scattered objects. The visualization results are shown in Fig. 10.

Table 5. Detection results on USAC-AOD dataset.

| Method | Plane | Car | mAP |
|---|---|---|---|
| YOLOv2 [10] | 96.60 | 79.20 | 87.90 |
| R-DFPN [51] | 95.60 | 82.50 | 89.20 |
| DRBox [52] | 94.90 | 85.00 | 89.95 |
| S$^2$ARN [53] | 97.60 | 92.20 | 94.90 |
| RetinaNet-H [28] | 97.34 | 93.60 | 95.47 |
| FADet [50] | 98.69 | 92.72 | 95.71 |
| R$^3$Det [28] | 98.20 | 94.14 | 96.17 |
| RSDet [31] | 98.04 | 94.97 | 96.50 |
| SCRDet++ [44] | 98.93 | 94.97 | 96.95 |
| PolarDet [47] | **99.08** | 94.96 | 97.02 |
| **ARP+R-EIoU (Ours)** | 98.91 | **95.17** | **97.04** |

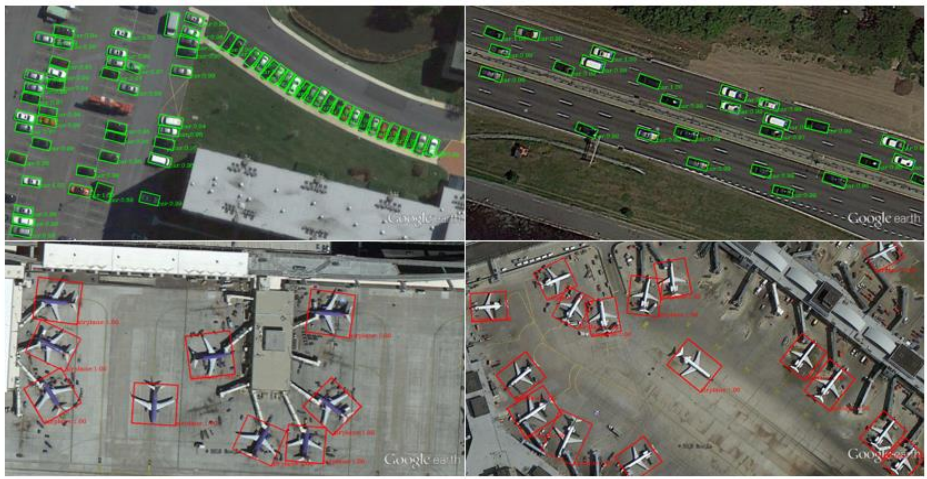

Figure 10. Visualization of detection results on UCAS-AOD dataset with our method.

Table 6. Detection results on ICDAR2015 dataset.

| Method | Recall | Precision | F-measure |
| --- | --- | --- | --- |
| RRPN [23] | 82.17 | 73.23 | 77.44 |
| RIL [36] | — | — | 77.6 |
| EAST [26] | 78.33 | 83.27 | 80.72 |
| TextBoxes++ [24] | 78.50 | 87.80 | 82.9 |
| RSDet [31] | — | — | 83.2 |
| R$^2$CNN [22] | 79.68 | 85.62 | 82.54 |
| DAL [46] | 80.5 | 84.4 | 82.4 |
| CSL [34] | 83.00 | 84.30 | 83.65 |
| RRD [25] | 80.00 | 88.00 | 83.80 |
| R$^3$Det [28] | **83.54** | 86.43 | 84.96 |
| DB [54] | 83.2 | **91.8** | **87.3** |
| **ARP+R-EIoU (Ours)** | 83.43 | 86.28 | 84.83 |

4.3.4 Results on ICDAR2015

The generalizability of our method across different scenes was also tested through experiments on the scene text detection dataset. As illustrated in Tab. 6, our algorithm yields an 84.83% F-measure, outperforming the majority of aerial detectors and text detectors, such as CSL [34], RRD [25]. However, our strategy does not account for the case of a significant number of lengthy texts that are incorrectly identified as multiple short paragraphs. As a result, our system still falls short of state-of-the-art scene text identification.

**5. Conclusion**

In this paper, the area ratio of parallelogram (ARP) representation method is suggested for the definition of oriented bounding boxes. It represents an oriented object by utilizing the central point coordinates, width and height of the smallest bounding rectangle of the oriented object and introducing three area ratios. Based on this, the area ratio of the oriented object and its smallest circumscribed rectangle is adopted to distinguish nearly horizontal objects. ARP method has a unique representation and facilitates offset learning. Thus, it is superior to the method of directly anticipating five-parameter or eight-parameter bounding boxes. Furthermore, R-EIoU loss is designed to connect the minimum circumscribed rectangle of the rotating object and three area ratios proposed by the ARP method. It accelerates network convergence and achieves higher performance in rotated detection when compared to the smooth L1 loss. Extensive detailed ablation experiments and comparative experiments on multiple rotation detection datasets, including DOTA, HRSC2016, UCAS-AOD, and ICDAR2015, proved the superiority of our method.

## Acknowledgments

We gratefully acknowledge the support of National Key R&D Program of China (2018YFB1308400) and Natural Science Foundation of Zhejiang Province (NO.LY21F030018).

Vision, pages 1440-1448, 2015.

[7] Shaoqing Ren, Kaiming He, Ross Girshick and Jian Sun. Faster r-cnn: towards real-time object detection with region proposal networks. IEEE Transactions on Pattern Analysis and Machine Intelligence. 39(6): 1137-1149, 2016.

[8] Kaiming He, Georgia Gkioxari, Piotr Dollár and Ross Girshick. Mask R-CNN. In Proceedings of the IEEE International Conference on Computer Vision, pages 2961-2969, 2017.

[9] Joseph Redmon, Santosh Divvala, Ross Girshick and Ali Farhadi. You only look once: unified, real-time object detection. In Proceedings of the IEEE Conference on Computer Vision and Pattern Recognition, pages: 779-788,2016.

[10] Joseph Redmon, Ali Farhadi. Yolo9000: better, faster, stronger. In Proceedings of the IEEE Conference on Computer Vision and Pattern Recognition, pages 7263-7271, 2017.

[11] Joseph Redmon, Ali Farhadi. Yolov3: An incremental improvement. arXiv preprint arXiv:1804.02767, 2018.

[12] Alexey Bochkovskiy, Chien-Yao Wang and Hong-Yuan Mark Liao. Yolov4: Optimal speed and accuracy of object detection. arXiv preprint arXiv:2004.10934, 2020.

[13] Wei Liu, Dragomir Anguelov, Dumitru Erhan, Christian Szegedy, Scott Reed, Chengyang Fu and Alexander C. Berg. SSD: Single shot multibox detector. In Proceedings of European Conference on Computer Vision, pages 21-37, 2016.

[14] Tsung-Yi Lin, Priya Goyal, Ross Girshick, Kaiming He and Piotr Dollar. Focal loss for dense object detection. In Proceedings of the IEEE International Conference on Computer Vision, Pages: 2980-2988. 2017.

[15] Hei Law, Jia Deng. Cornernet: Detecting objects as paired keypoints. In Proceedings of European Conference on Computer Vision, pages 734-750, 2018.

[16] Kaiwen Duan, Song Bai, Lingxi Xie, Honggang Qi, Qingming Huang and Qi Tian. CenterNet: Keypoint triplets for object detection. In Proceedings of the International Conference on Computer Vision, pages 6568–6577, 2019.

[17] Peng Sun, Yongbin Zheng, Zongtan Zhou, Wanying Xu, and Qiang Ren. 2020. R4Det: Refined single-stage detector with feature recursion and refinement for rotating object detection in aerial images. Image and Vision Computing 103:104036.

[18] Jiahui Yu, Yuning Jiang, Zhangyang Wang, Zhimin Cao and Thomas Huang. Unitbox: An